# Deep Learning Models to Study Sentence Comprehension in the Human Brain

Sophie Arana[†,1,2,3,*], Jacques Pesnot Lerousseau[†,1] and Peter Hagoort[2,3]

[1] Department of Experimental Psychology, University of Oxford, OX2 6HG, United Kingdom

[2] Max Planck Institute for Psycholinguistics, 6525 XD Nijmegen, The Netherlands

[3] Donders Institute for Cognition, Brain and Behaviour, Radboud University, 6525 AJ Nijmegen, The Netherlands

* Correspondence: arana.sophie@gmail.com

[†] Co-first authorship

**Disclosure of interest:** The authors declare no competing interests.

**Acknowledgements**: We thank Christopher Summerfield, Micha Heilbron and Jessica A. F. Thompson for their precious comments.

**Funding sources:** Work supported by Fondation Pour l'Audition (J.P.L.)

**Author contributions:** Conceptualization S.A., J.P.L. and P.H.; Supervision P.H.; Writing – original draft S.A. and J.P.L.; Writing – review & editing S.A., J.P.L. and P.H.

# Abstract

Recent artificial neural networks that process natural language achieve unprecedented performance in tasks requiring sentence-level understanding. As such, they could be interesting models of the integration of linguistic information in the human brain. We review works that compare these artificial language models with human brain activity and we assess the extent to which this approach has improved our understanding of the neural processes involved in natural language comprehension. Two main results emerge. First, the neural representation of word meaning aligns with the context-dependent, dense word vectors used by the artificial neural networks. Second, the processing hierarchy that emerges within artificial neural networks broadly matches the brain, but is surprisingly inconsistent across studies. We discuss current challenges in establishing artificial neural networks as process models of natural language comprehension. We suggest exploiting the highly structured representational geometry of artificial neural networks when mapping representations to brain data.

# I. New analytic tools have enabled the study of brain activity during sentence comprehension.

Recent advances in natural language processing, *i.e.* the automatic analysis of natural language by computer algorithms, have greatly impacted the study of brain activity during sentence comprehension. Two main innovations are at the core of these successes: representing words as low-dimensional dense vectors and learning such representations with artificial neural networks trained on large text corpora (Goldberg, 2016).

### a. Learning word vectors with artificial neural networks

In order to build natural language processing algorithms, it is necessary to represent language units, such as words, as numbers. This operation is called embedding. To illustrate this point, let's take a toy language composed of five words: *King, Queen, Man, Woman, Men*. A naive embedding would be to convert each word as a one-hot vector, where each word is its own dimension: *King* is the first dimension [1, 0, 0, 0, 0], *Queen* is the second dimension [0, 1, 0, 0, 0], *Man* is the third dimension [0, 0, 1, 0, 0], and so on. Such an embedding has two disadvantages. First, it is memory inefficient, as the vector size increases quickly with the size of the vocabulary: in our example, n dimensions encode at most n words. Second, there is no similarity structure between the words as each word is completely independent from the others, which is a problem for generalisation purposes, such as learning a new word and quickly inferring its meaning based on similar words in the vocabulary.

A better implementation is to use distributed encoding (Harris, 1954; Hinton, 1986; Rumelhart, Hinton, & Williams, 1986), *i.e.* to construct the representation of each word as a set of multiple features. Usually distributed encoding is thought to represent semantic features rather than visual, orthographic, auditory or phonological properties of words, although in principle any feature could be represented through a distributed code. In our toy example, one can represent each word as a 3-dimensional vector with "gender" as the first dimension (0: male, 1: female), "number" as the second dimension (0: singular; 1: plural) and "regal" as the third dimension: *King* is [0, 0, 1], *Queen* is [1, 0, 1], *Male* is [0, 0, 0], *Woman* is [1, 0, 0] and *Men* is [0, 1, 0]. Such a distributed representation has multiple advantages. First, it is memory efficient: in our example, n dimensions encode at most $2^n$ words. Its dimensionality is thus much lower than the naive embedding one, hence the term "low-dimensional". In practice, the number of dimensions in recent models ranges from about 50 to a few hundred. Second, it allows performing semantically meaningful operations on vectors (Mikolov, Chen, Corrado, & Dean, 2013): the operation *King* + *Woman* = [0, 0, 1] + [1, 0, 0] = [1, 0, 1] = *Queen*. Third, it facilitates generalisation by reusing the same features to encode new words and thus infer their similarity to the other words: in our toy language, the new word *Kings* would be encoded as [0, 1, 1], which happens to be close to *King* ([0, 0, 1]) and far from *Woman* ([1, 0, 0]). On a side note, this inference will only be meaningful if the initial distributed representation is meaningful. For example here, "gender" as an input variable has been imposed by hand.

Moving from representing each word as a unique dimension to representing them instead as dense vectors has been described as "perhaps the biggest conceptual jump" in natural language processing (Goldberg, 2016). Nowadays, nearly all tools developed in natural language processing use such distributed vector representations of individual words.

Recent embeddings typically consist of several hundreds of dimensions, for example 640 in (Mikolov et al., 2013) or 512 in (Vaswani, Shazeer, & Parmar, 2017). Contrary to our toy example, the dimensions of these embeddings are not easily interpretable as they do not correspond to salient features, such as gender.

*Learning word embedding by training artificial neural networks on large text corpora.*

Finding an optimal low-dimensional word embedding is a difficult problem, especially when the vectors have several hundreds of dimensions. The current approach in natural language processing is to use large text corpora to learn the appropriate embeddings. This relies on the distributional hypothesis: the fact that semantic features that distinguish the meanings of words are reflected in the statistics of their use within large text corpora (Firth, 1957; Mikolov et al., 2013; Mitchell et al., 2008), *i.e.* that words that have a close semantic proximity tend to appear within similar contexts. This means that the embeddings do not use the statistics of the world nor the co-occurrence of words and real-world situations directly but only through the statistics internal to the text corpora. Learning these statistics is typically done by artificial neural networks and deep learning techniques (LeCun, Bengio, & Hinton, 2015). These artificial neural networks are composed of artificial neurons that realise simple operations on inputs. The artificial neurons are organised in layers. In recurrent architectures, the output of each layer is fed back as input to the same layer at the next time step. In serial architectures, such as transformers, the output of the artificial neurons of one layer are the input to the artificial neurons of the next layer. Note that "layer" does not refer to the six cortical layers of the brain but rather to the stage of an artificial neuron in this series of pools of artificial neurons. Two main artificial neural network architectures dominate the field: recurrent neural networks and transformers.

First, recurrent neural networks have been extensively used in natural language processing. They process input sequences one element at a time, while maintaining a separate state vector that contains the information about previous context. The most used architecture is the long-short term memory unit (LSTM) (Mikolov, Karafiát, Burget, Černocký, & Khudanpur, 2010; Sundermeyer, Schlüter, & Ney, 2012). In such architectures, the input layer is a one-hot encoding vector that corresponds to the word that is presented. The next layers are composed of LSTM units. Each LSTM unit is itself composed of three gates that control which information of the hidden state vector is going to be forgotten, maintained and integrated to the input (Hochreiter & Schmidhuber, 1997). This allows for flexible gating of information to either maintain it over a long period of time or quickly forget it when necessary. Over training, the network learns to build and maintain the best possible state vector, *i.e.* the vector that captures the most information about the sentence. Depending on the task it has been trained on, the network can output next word predictions, next sentence selections and sentence topic predictions (Ghosh et al., 2016).

Second, the transformer architecture (Vaswani et al., 2017) is currently the state of the art for natural language processing. Contrary to recurrent neural networks that process sequences one element at a time, transformers process whole sequences in parallel. In the input layer, each word of the sentence is fed to the network as a one-hot vector. Each word is then independently embedded by a first embedding layer. This produces a context-free embedding of the words. A positional encoding vector is then added to the context-free embedding of each word so that the model has information about the word order. Then, transformers consist of a series of embedding layers, where the embedding of each word is mixed with the embeddings of the surrounding words to produce a context-dependent embedding. As a result, the first embedding layer is a context-free embedding of each word while the last embedding layer provides a context-dependent embedding of each word. The central component of each embedding layer is the "attention head" that controls the amount of mixing between word embeddings and the precise locus of this mixing through a "self-attention" mechanism. It should be noted that "attention" only refers vaguely to the usual notion of attention in psychology and the reader should treat them as two different concepts. More specifically, an attention head proceeds in three steps. (1) First, the attention head computes three projections of each word, called the "query", "key" and "value" vectors. This projection is learned by the model during training. (2) Second, for each word, the attention head computes a similarity score between the "query" vector of the word and the "key" vectors of all words, by taking the dot product of each vector pair. This gives a value that indicates to what extent the "key" of each word matches the "query" of each word. (3) Third, the attention head computes a weighted average of the "value" vectors for each word, where the weight is equal to the similarity score computed above. This weighted average is the new embedding of each word, ready to be passed to the next layer for another round of mixing.

To illustrate further, let's take the example sentence "*I put my cash in the bank*". The interest of this example is the polysemic word "*bank*", which can refer to a financial institution or to land alongside a river. To disambiguate its meaning, an attention head might produce a "query" vector for "*bank*" that has high values for features like "*institution-related*", "*money-related*", "*river-related*", "*lake-related*". (in plain text, the attention head asks: *does anyone in the sentence know if we are talking about money-related stuff or about river-related stuff?*). The "key" vector of "*cash*" will probably have a high similarity score with the "query" vector because it has a high "*money-related*" feature value (in plain text, the word "*cash*" responds: *yes, I have information to know that we are talking about money-related stuff and not river-related stuff!*). The attention head will finally mix the embedding of the word "*bank*" with the embedding of the word "*cash*". As a result, the "*river-related*" feature disappears and the "*money-related*" feature is reinforced in the context-dependent embedding of the word "*bank*". To sum up, one can think of an attention head as a series of computations that project the words in a subspace, measure the similarity between each word in that subspace, and mix the representation of the words according to this similarity measure. In practice, the features of the word vector are not built-in but rather learned through training, and contrary to our example they are usually not easily determined in any interpretable way. Further, transformers are composed of multiple attention heads in each layer and also involve further mixing strategies. Transformers are usually trained to predict a masked word in a sentence or a paragraph. The most common transformer models that have been used in neuroscience are BERT (Devlin, Chang, Lee, & Toutanova, 2018), GPT-2 (Radford, Wu, Child, Luan, & Amodei, 2019) and GPT-3 (Brown, Mann, & Ryder, 2020).

Learning word embeddings on large text corpora by means of artificial neural networks, has produced excellent results in a variety of natural language processing tasks, such as question answering, translation or text summarisation. Further, this approach has provided new models of language processing for cognitive neuroscientists interested in how the brain processes language. Indeed, these models have two advantages. First, they are usually trained end-to-end with almost no a priori knowledge on how language works. This is an advantage because they do not depend on linguistic theories and thus provide more objective measures on language. Further, this releases additional assumptions concerning the fact that humans are born with in-built knowledge of the language structure. Second, they can process natural language, thus providing tools to analyse human behaviour and brain activity during naturalistic tasks. As a consequence, a multitude of methods to compare the brain with these language models have been developed.

### b. Comparing brain activity with artificial neural networks

The most common paradigm involves presenting the same words or sentences to human participants and to the models. Three approaches have been used to then compare human behaviour and brain activity with the models: directly correlating brain activity with the model's activity, comparing behaviour and brain activity with metrics derived from the model's outputs, and comparing the geometry of the representations extracted from the brain activity and the model's activity. Each approach involves different assumptions and can lead to different conclusions concerning the link between the brain activity and the models.

*Comparison of brain activity with the model's activity.*

The first approach is to directly compare the patterns of brain activity during the presentation of words to the word embeddings of the model. The assumption is that if the brain is using the same embedding strategy as the model, then one should be able to build a mapping between the brain activity and the word embeddings. Usually, a further assumption is that this mapping is linear and can thus be approximated by a linear regression.

The linear regression approach was employed for the first time by Mitchell and collaborators in 2008 (Mitchell et al., 2008). The authors measured brain activity during the presentation of single words using fMRI. The activity in each voxel was then decomposed as a weighted sum of the corresponding word embedding. The mapping between the word embedding and each voxel was learned by a linear regression and was used to predict brain responses to words that were not used during training. Geometrically, a linear regression can be thought of as a hyperplane. In this context, the hyperplane is a tilted flat surface in the space whose dimensions correspond to the embedding space dimensions and a dimension corresponding to the brain response. The brain response associated with a given word can be described by a projection of the word in the embedding space on a line in the tilted direction of the hyperplane. Therefore, the success of the linear regression in one voxel implies that the activity in this voxel is correlated to the projection of the words onto a line in the embedding space. This linear mapping approach can be equally applied at the sentence level by replacing the word embedding with an embeddings vector for the integrated sentence meaning. Using sentence-level embeddings, recent works have extended this approach to other regression techniques, such as linear ridge regression (Caucheteux & King, 2022), other models, such as GPT-2 (Goldstein, Dabush, et al., 2022), and other brain recording modalities, such as MEG (Wehbe, Vaswani, Knight, & Mitchell, 2014) or ECoG (Goldstein, Zada, et al., 2022). These works are reviewed in detail in the next part.

*Inferring identity between brain and models.*

Not every comparison method is well suited to establish correspondence between a model and the brain. A single correlation between one model and the brain data, as is typically done using the regression approach, provides limited information. Indeed, the level of explanation at which one can infer identity based on model fits is difficult, because similar functions do not automatically imply similar realisations (Guest & Martin, 2021; Jonas & Kording, 2017). A cautious approach is usually to refrain from inferences concerning explanation at the lowest levels, like those about detailed implementation. Further, inferences are most meaningful when based on a contrastive approach, comparing model fits between multiple models that differ in one aspect.

*Overfitting.*

Any work that compares large artificial neural networks and brain data using regression techniques faces the issue of overfitting. This is partly because the number of trials used as training data for the regression is usually much smaller than the number of parameters to train. In these conditions, even a bad model could learn the idiosyncratic characteristics of the training set and artificially fit the training data. To reduce this issue, standard methodological tools are used. First, regularisation methods, such as ridge regression (Tikhonov, 1963) or lasso regression (Tibshirani, 1996), penalise the regression for large parameters. This biases the regression towards sparse sets of parameters, *i.e.* sets with a low number of non-zero parameters. Such sets are less prone to overfitting. Second, cross-validation reduces overfitting by using independent datasets for training and testing the models fit to the brain.

*Comparison of brain activity with metrics derived from the model.*

The second approach is to compare brain activity with metrics derived from the model's outputs. The main assumption is that if the brain and the models share similar computations, they will produce similar outputs. This does not assume that the brain and the model rely on similar representations, but rather that they produce the same outputs or that they converge toward the same result. The most common method is to compare the surprise of the model and the brain activity. Surprise is a quantity developed in the field of information theory (Shannon, 1948) to measure the degree of unexpectedness of a stimulus. It has a precise mathematical formulation: $-\log_2 P(w)$ where $P(w)$ is the probability of occurrence of the word w. It nonetheless corresponds to the intuitive notion of surprise, *i.e.* it is low when a word is highly expected by the model and high when a word is highly unexpected by the model. A further usual assumption is that the brain activity scales with the surprise, such that highly surprising words elicit large brain activity while highly unsurprising words elicit small brain activity (Mars et al., 2008).

Information-based surprise has been used in a recent paper by Heilbron and collaborators (Heilbron, Armeni, Schoffelen, Hagoort, & de Lange, 2020). The authors compute the surprise of a language model, GPT-2, relative to different linguistic levels, namely syntactic, phonemic and semantic. Syntactic surprise refers to the extent to which the model expects the syntactic category of each word. Similarly, phonemic and semantic surprise correspond to the extent to which the model expects the phonemes and the semantic features respectively. The authors then correlate the brain activity measured by EEG and MEG with the surprise computed from the model with respect to the different linguistic levels to reveal the location and time windows during which the brain processes each linguistic level. Other recent works have used this method with other brain imaging, such as fMRI (Brennan, Dyer, Kuncoro, & Hale, 2020; Schmitt et al., 2021a), and with other output measures, such as entropy (Donhauser & Baillet, 2020).

Another related form of surprisal that has been shown to modulate the brain's responses is Bayesian surprise., *i.e.* how much a model's belief changes based on an incoming word (Itti & Baldi, 2009). In the context of sentence comprehension, this has been demonstrated for example by Rabovsky et al's sentence gestalt model, a recurrent neural network architecture trained on sentence comprehension. The word-induced update in the hidden unit activations of the network came to reflect the update of a probabilistic sentence representation. The magnitude of this update was then shown to predict amplitude modulations in neural activity across tasks (Rabovsky, Hansen, & McClelland, 2018). To our knowledge the Bayesian surprisal metric has not yet been used in conjunction with large-scale language models as a comparison for human brain data (but see M. Kumar et al., 2022 for predicting human narrative segmentation).

*Comparison of the geometry of the representations.*

The third approach is to compare the geometry of the representations of the words computed from the model and from the brain activity. Unlike the two previous approaches, this method does not necessarily assume a linear relationship between the brain and the model. Geometry is usually inferred through representational similarity analysis (Diedrichsen & Kriegeskorte, 2017; Kriegeskorte, Mur, & Bandettini, 2008), which consists in computing the matrix of dissimilarities between activity patterns elicited by each word. A matrix of dissimilarities is obtained for the brain activity and for the model. The test then consists in comparing both matrices, usually by calculating rank-based correlations.

This approach has been used to compare the evolution of the semantic representation during the sentence presentation in both a BERT model and brain activity recorded by EEG and MEG (Lyu, Tyler, Fang, & Marslen-Wilson, 2021). Overall, it has been less used to study sentence comprehension compared to the two previous ones. This is surprising given that, at first glance, the estimation of the representational geometries is less computationally demanding and requires less assumptions about the distribution of the data than the linear encoding approach (Diedrichsen & Kriegeskorte, 2017). One reason for the lack of research using representational similarity analysis is the difficulty to build a reliable dissimilarity matrix in the context of natural language comprehension. In natural language, most of the words are not repeated over the course of a text. This means that most words correspond to unique trials, leading to very large and noisy dissimilarity matrices. Nonetheless, it should be noted that, in practice, linear encoding models and representational similarity analysis leads to similar results (Thirion, Pedregosa, & Eickenberg, 2015).

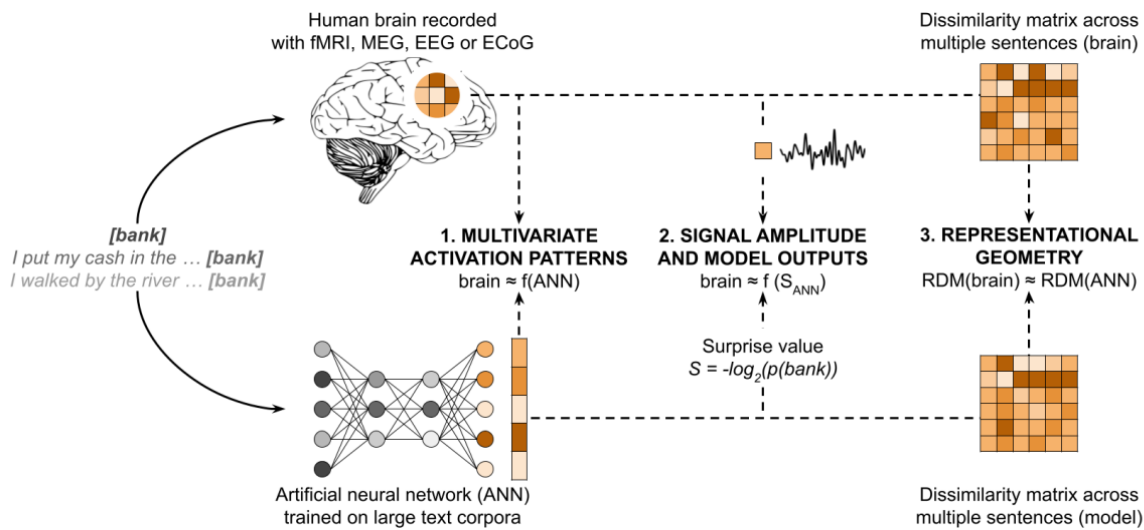

**Figure 1. Overview of the modelling and multivariate analytic tools to study brain activity during speech comprehension.** Brain activity is recorded during the presentation of single words or sentences. Artificial neural networks trained on large text corpora, such as LSTM or transformer networks, are used as models of natural language processing: their activity is also recorded during the presentation of the same single words or sentences. Three approaches have been used to then compare brain activity with the models. **1.** Directly correlate brain activity with the model's activity, for example with linear (ridge) regression. **2.** Compare behaviour and brain activity with metrics derived from the model's outputs, such as surprise values. **3.** Compare the geometry of the representations extracted from the brain activity and the model's activity by comparing the representational dissimilarity matrices (RDM).

*Mapping detailed model embeddings onto coarse brain measures.*

Artificial neural networks usually learn separate embeddings for each word in the sentence rather than one fixed size sentence embedding. This causes a challenge whenever a mapping onto a modality with coarser resolution is attempted. For example, the hemodynamic response recorded in fMRI data unfolds over several seconds and will inevitably span multiple words given naturalistic stimulus presentation. There are multiple ways to circumvent this issue.

When mapping language models onto fMRI data, most researchers have relied on averaging or concatenating word embeddings to match the resolution of the neural data (Anderson et al., 2021; Toneva, Mitchell, & Wehbe, 2020, 2022a; Wehbe, Vaswani, et al., 2014). In recent transformer models, contextualised embeddings aggregate information on preceding word meaning, which makes it possible to simply rely on individual embeddings from the final hidden layer, such as the sentence final word (Schrimpf et al., 2021). Another approach specific to pretrained BERT transformers is to use an additional token embedding as an aggregate sentence embedding. BERT transformer models currently available are commonly trained to infer whether one sentence is a plausible continuation of another. This sentence classification task requires a pair of sentences as input to the model. To indicate the separation of a potential second sentence in the input, BERT requires each input sentence to begin with the meaningless special token [CLS]. Just like every word, the embedding at the special token [CLS] position is getting increasingly mixed with the embeddings of the subsequent words in higher-level layers of the model. As a result, the embedding at the [CLS] position reflects an aggregate of the embedding of the words that constitutes the sentence. Both the [CLS] token's embedding or an aggregate measure of all word embeddings through max-pooling have been used to model human data on sentence-level classification tasks such as judging semantic textual similarity or sentiment analysis (Devlin et al., 2018; Reimers & Gurevych, 2019). However, token-specific embeddings often do not reach maximal performance without further fine-tuning (Devlin et al., 2018; Reimers & Gurevych, 2019). Overall, the choice of how to integrate sentence embeddings is rarely explicitly motivated and only few papers have directly compared how different approaches affect a model's fit to brain data.

# II. The brain represents words as low-dimensional dense vectors in a context-dependent fashion.

The first contribution of the combined development of models of natural language processing and methods to compare the brain activity with these models is a precise investigation of the representational code of the words' meaning during sentence comprehension.

### a. Word meaning representation as low-dimensional dense vectors

Hypotheses about the neural representation of words can be placed on a spectrum. On one side, the word representations could be localised, with each word being encoded by a specific population of neurons and no overlap between different words. On the other side, the representation of the words could be distributed, with all words being encoded in the same population of neurons but with different patterns for each word. This second hypothesis has received a lot of attention, notably because of the success of the distributed representation in the field of natural language processing. As a consequence, a lot of work has been done to compare the dense vector representation used in natural language processing to the brain activity.

As previously cited, the work of Mitchell and collaborators in 2008 (Mitchell et al., 2008) was the first to test the distributed representation hypothesis in the human brain. In this study, the authors predicted fMRI activity using a linear combination of the vector representation of the presented words. A strong implication of a successful linear mapping is that if the weights of the linear regression are diverse enough across all voxels and if the number of voxels is large enough, the brain activity potentially spans the entire embedding space of the model. This would imply that the brain represented words in a similar fashion as the model, *i.e.* as low-dimensional dense vectors distributed in space. Such a representation would allow for interpolation, which is incompatible with a naive embedding, with one feature for every word. However, note that this implication is only probable and not guaranteed. Indeed, if the words in the embedding space are linearly separable from one another, a similar approach would succeed even if the brain used the naive embedding, with one feature for each word. Nonetheless, this scenario is improbable, because there are many more words than dimensions in the embedding space and thus many more configurations where most words are not linearly separable than configurations where they are.

Following this proof of concept, this result has been replicated and extended in later works. First, this result has been replicated using complete sentences rather than single words (Anderson et al., 2019; Huth, de Heer, Griffiths, Theunissen, & Gallant, 2016; Millet & King, 2021; Millet et al., 2022; Pereira et al., 2018; Wehbe, Murphy, et al., 2014). Second, the distributed representation has been shown to be at least partly stable across individuals, as the spatial localisation of the linear mapping between the word embedding and the fMRI data appears to be somewhat overlapping across individuals (Huth et al., 2016). Third, it has been shown that the distributed representation of the semantic information is encoded in different cortical areas and is independent from non-semantic features of the words. During visual stimulation, the encoding is independent from the visual appearance of the words (Wehbe, Murphy, et al., 2014). Similarly, during auditory stimulation, the encoding is independent from the spectral and articulatory features of words (de Heer, Huth, Griffiths, Gallant, & Theunissen, 2017). The encoding is also independent from the syntactic role of a word in the sentence (Wehbe, Murphy, et al., 2014). Indeed, multiple cortical regions represent semantic information about the words irrespective of their grammatical role, such as grammatical subject or object (Anderson et al., 2019). Finally, the encoding of semantic information has been proposed to be partly modality independent, as the geometry of representation of visually presented words was correlated to those of auditorily presented words in the left pars triangularis (Liuzzi et al., 2017). Recently, Popham and collaborators have shown that the semantic representations of images and speech are aligned at the border of the visual cortex (Popham et al., 2021).

Most of these studies used linear regression to compare fMRI single voxel activity to the vector representation of the words. However, this minimally proves that at least one direction in the embedding space is encoded in the brain. One could argue that the brain may represent the language model's embedding space only partly but not in its entirety. In the most extreme scenario, one could even argue that the voxels are actually all responding to the same single direction in the embedding space, and thus not at all making use of a dense low-dimensional representation of the words. In this case, an encoding model would have good performances at predicting brain activity from the word embeddings but a decoding model would have poor performances at predicting the word embeddings from brain activity. To address this criticism, Goldstein and collaborators have successfully shown that information in the word embedding can be used to predict brain activity but also that brain activity can be used to predict the word embeddings (Goldstein, Dabush, et al., 2022; Goldstein, Zada, et al., 2022). The linear mapping between the brain and the neural network performed relatively well in both directions, with a cross-validated correlation score of 0.15 for the encoding and a prediction score of 0.70 for the decoding in the IFG. The good performances in the two directions with the same linear model suggest that the brain representations are not restricted to a small subset of the embedding space used by the natural language processing models. Indeed, if the geometric relationships between words were only very partly overlapping between the brain and the neural networks, such good cross-validated performances would be unlikely. Another approach put forward to overcome the pitfalls of a unidirectional correlation measure when comparing models and brains is the "direct interface" test, which consists of evaluating model performance after directly using the mapped brain activity in place of a single layer activation in the neural network. This approach additionally ensures that any similarity in encoding between brain and networks are functionally relevant and not just driven by spurious variance in the signal. It has been used in the context of object recognition in vision (Sexton & Love, 2022), but never in language comprehension.

### b. Word meaning representation in context

The previous part focused on models that use context-free embeddings, *i.e.* fixed word vectors that do not change depending on the surrounding context. However, context helps to disambiguate the meaning of words and to assign them the correct semantic features. The polysemic word "*bank*" refers to a financial institution in the context A: "*I put my cash in the bank*", but it refers to a landscape in context B: "*I walked by the river bank*". Consequently, the word vector representation of the word "*bank*" should not be the same in the two contexts. The semantic features of "*bank*" in context A should be related to money and finance, while the semantic features of "*bank*" in context B should be related to river and walking. Even for words with unambiguous word sense, context often directs attention to a subset of semantic features of a word's meaning. For example, in a phrase like "*throwing a banana*", features relating to the colour or taste of a banana are less important than its shape or weight.

The first study using contextualised word embedding was conducted by Wehbe and collaborators (2014). The authors used a recurrent neural architecture to generate two vectors for each word of the sentence, one representation of the context and one contextualised embedding of the word. Based on these two vectors they predicted MEG activity before, during and after the presentation of each word using a linear regression. Their results show that before the presentation of a word in a sentence context, the context representation predicts MEG activity. Shortly after the presentation of the word, the contextualised word representation predicts MEG activity. Finally, after the presentation of the word, the updated representation of the context predicts MEG activity. This can be interpreted as evidence that the brain represents the context as a latent variable, and integrates the meaning of the word to this context representation to produce a context-dependent embedding (Wehbe, Vaswani, et al., 2014).

Further, it has been shown and replicated multiple times that contextualised embeddings are better predictors of brain activity than context-free ones in large parts of the language network. This effect has been shown in EEG and MEG recordings during the presentation of sentences with a verb-object noun relationship (Lyu et al., 2019) and in fMRI and MEG recordings during the presentation of narratives (Caucheteux & King, 2022; Toneva & Wehbe, 2019). In a recent article, Schrimpf and collaborators compared a large number of models, including models that take context into account or not, to brain data recorded during the presentation of narratives. Contextual models were systematically associated with better fit to the brain data compared to context-free models, for different datasets and different recording modalities, including fMRI, MEG and ECoG (Schrimpf et al., 2021). One surprising result is that despite the apparent similarity between the brain and LSTM neural networks in terms of stimulus presentation (serial presentation) and computations (recurrent neural architectures), transformers are better predictors of brain activity. More generally, the performance of the neural networks in language tasks is a key predictor of their ability to match brain activity (Caucheteux & King, 2022; Schrimpf et al., 2021). Predictability of brain activity has been interpreted by some as suggesting a "convergence" in processes between models and the brain. This inference has also been criticised, however, because measures of predictive power are often averaged across inputs that vary in a large number of feature dimensions, therefore limiting its explanatory power (Bowers et al., 2022).

What explains the better fit of the contextualised representation? The advantage of contextualised representation could be explained by contextual information improving word embeddings, for example because polysemic words are disambiguated. It could also be explained by the fact that the contextualised embedding contains information about previous words which could be used to account for variance in brain activity related to those previous words. When systematically varying the amount of context that is used by the recurrent language model, the better fit of the contextualised representation to the brain was reported to be due to both of those factors simultaneously (Jain & Huth, 2018). Also, words presented with more context were associated with stronger fMRI activity and better encoding (Deniz, Tseng, Wehbe, & Gallant, 2021).

Finally, several artificial neural networks have been shown to capture sentence meaning beyond a simple "sum of its parts". As models of brain activity, they outperform alternative models that rely on simpler operations such as point-wise averages or concatenation of context-free word embeddings, especially when distinguishing neural representations evoked by closely related sentences (Anderson et al., 2021; Sun, Wang, Zhang, & Zong, 2019). Moreover, the contextualised embeddings of recent language models seem to capture semantic information beyond individual word meaning, emerging through coactivation with a combination of words, such as activating aspects of the verb "*reading*" when hearing the phrase "*the girl began the book*" or "*eating*" for the phrase "*the goat finished the book*" (Toneva, Mitchell, & Wehbe, 2022b).

# III. The processing hierarchy within artificial and neural language models

It appears from the literature summarised in the previous part that words are encoded in distributed neural activation patterns that are well captured by artificial language models. However, the mere mapping between natural language processing models and neural representations is not considered the end goal of this line of research. Through such a mapping, the hierarchical architecture of artificial neural networks can reveal a parallel functional hierarchy in the brain, similar to what has been demonstrated for the domain of vision (Cichy, Khosla, Pantazis, Torralba, & Oliva, 2016; Eickenberg, Gramfort, Varoquaux, & Thirion, 2017; Güçlü & van Gerven, 2015).

### a. Contrastive approach to localise brain networks involved in contextualisation processes

From decades of neuroimaging research using mainly univariate analysis methods the language community has collected a wealth of data and formulated detailed hypotheses about the functional involvement of brain regions during sentence processing. We know that the left posterior and middle temporal cortex activates early after word onset and in a modality-independent manner (Arana, Marquand, Hultén, Hagoort, & Schoffelen, 2020) with its activation being strongly time-locked (Hultén, Schoffelen, Uddén, Lam, & Hagoort, 2019). This pattern suggests a largely lexicalized, bottom-up process and hence this area is thought to encode lexical representations including lexicalized syntactic representations in posterior parts (Matchin, Brodbeck, Hammerly, & Lau, 2019; Nelson et al., 2017). More anterior regions of the cortex, including the anterior temporal lobe as well as the inferior frontal gyrus, are usually activated only subsequently but sustain activation for longer (Arana et al., 2020); (Brodbeck, Presacco, & Simon, 2018). Moreover, the inferior frontal gyrus has been shown to engage in functional coupling with the temporal cortex, likely enabling integration of multiple words over longer time scales (Baggio & Hagoort, 2011; Hultén et al., 2019; Schoffelen et al., 2017). This network of frontal and posterior temporal regions is also consistently activated during tasks requiring semantic control (Solomon & Thompson-Schill, 2020), with damage to both regions leading to similarly impaired semantic access in a variety of tasks including conceptual combination (Jackson, 2021; Jefferies & Lambon Ralph, 2006; Jefferies, 2013). The anterior temporal lobe (ATL) has been suggested to be sensitive to both syntactic and semantic features of a sentence (Rogalsky & Hickok, 2009) and to play a role in conceptual combination (Pylkkänen, 2019; Zhang & Pylkkänen, 2015). In studies contrasting lists, phrases and sentences, graded effects in ATL have been reported in response to increasingly larger contextual units (Matchin, Brodbeck, et al., 2019). At the same time, evidence from neuropsychological disorders and simulation studies suggest a causal role for ATL already at the single-word level, specifically for tasks requiring semantic memory such as picture naming (Lambon Ralph, Jefferies, Patterson, & Rogers, 2017; Shimotake et al., 2015). ATL activation is thus unlikely to be restricted to combinatorial processing. Attempts to reconcile the discrepant findings regarding ATL function have pointed out a common sensitivity to conceptual specificity during both single and multi word processing (Westerlund & Pylkkänen, 2014; Zhang & Pylkkänen, 2015). Finally, the temporal parietal junction, including angular gyrus, increases in activation at later time points in the sentence (Matchin, Brodbeck, et al., 2019). Although the functional role of angular gyrus in sentence processing is still debated, there is now accumulating evidence for its involvement in higher-level event-related processing (Binder & Desai, 2011; Branzi, Pobric, Jung, & Lambon Ralph, 2021; Leonardelli & Fairhall, 2022; Matchin, Liao, Gaston, & Lau, 2019).

The univariate results summarised above are based on either contrastive designs such as comparing lists and sentences, minimal word pairs with compositional and non compositional meaning or on regressors that quantify structural complexity according to linguistic parsing models. In contrast, deep neural networks can model hierarchical processing stages based on naturalistic language data without the need to rely on engineered task designs. Their hierarchical architecture is well-suited for capturing the gradient of contextualization seen in the brain. Exploring the hierarchical gradients that emerge in artificial language models may hence provide complementary insights into the computational hierarchy of the brain.

### b. A gradient of contextualisation

As detailed in part II, context-dependent representations of the words have been shown to better model human brain data than context-free word embeddings. Moreover, studies using the linear mapping approach in a spatially resolved manner have shown that different timescales of contextualization may be associated with neural activity in distinct brain areas. This approach involves computing the voxel-wise fit for each voxel individually or group of voxels and comparing the fit across several language models that encode varying degrees of context. While multiple studies report a spatially varying gradient of contextualization in the brain (Antonello, Turek, & Vo, 2021; Jain & Huth, 2018; Qian, Qiu, & Huang, 2016; Schmitt et al., 2021b; Toneva & Wehbe, 2019), they vary in terms of the exact mapping of this gradient onto brain areas. While some find highly contextualised representations to be restricted to anterior temporal lobe and posterior temporal lobe (Toneva et al., 2022a), other findings suggest an inferior-superior gradient spanning temporal and inferior parietal cortex, emphasising angular gyrus and precuneus as processing information at longer contextual windows (Antonello et al., 2021; Schmitt et al., 2021b). Again others suggest a combination of all of the above (Caucheteux & King, 2022).

Even though the deep learning models have the means to apply continuous measures of contextual richness, their conclusions with respect to linguistic function remain rather coarse in comparison to the collective insights stemming from univariate studies. The main distinction relies on a binary split into brain areas preferring short or no context and brain areas preferring long contexts. It remains to be explored whether the observed differences in neural representations are sufficiently explained by this binary categorisation into non-contextual and contextual meaning representations or whether we can identify intermediate stages of context-dependence. For example, can we identify meaningful definitions of distinct 'chunks' of context, other than the amount of characters that can distinguish sentences, paragraphs and narratives?

On top of a rather coarse localisation, the internal discrepancy in the findings makes it hard to reconcile with the previous literature. Namely, studies differ on the precise brain areas that they report, spanning all four temporal, parietal, frontal and occipital lobes. Several factors can explain the discrepancies in the reported brain localisation:

First, the studies use different neuroimaging methods that have different profiles of spatial and temporal resolution. For example, MEG has millisecond temporal resolution, which is excellent to differentiate the neural response to consecutive words, but it has a relatively poor spatial resolution, which makes it inappropriate to differentiate activity in neighbouring neural populations. On the contrary, fMRI has a poor temporal resolution but millimetre spatial resolution, which makes it less appropriate to differentiate activity of words presented close in time, as previously stated.

Second, the fact that successful encoding of brain activity has been observed in different brain areas could indicate that the neural code for word meaning is partly redundant or at least correlated across brain areas. In a neural network, the activity across successive layers is correlated because the layers receive input from one another. Similarly, correlations between brain areas could emerge from the functional connectivity between regions, the activity of one area being the input of another, or from the fact that two areas receive the same inputs but perform different computations.

Finally, the discrepancies could emerge from the variety of artificial neural networks that have been used to study brain activity. These models vary in architecture, training data, and objective function. It is thus unclear to what extent these different models produce different word embeddings, and whether these differences are important when fitting brain activity. On the one hand, the brain could represent a common part of the information shared across all models. On the other hand, some models might be better than others and this might depend on the brain areas. For example, it has been shown that the geometry of word meaning representations is not the same if it is computed from behavioural ratings of semantic features or from statistical information about its linguistic contexts. This difference is reflected in the brain regions that encode each representation (Wang et al., 2018). To answer these questions, a recent paper suggests that the variety of word embeddings used by the different neural networks actually span a meta-embedding space, whose main axes correspond to how much and how contextual information is taken into account (Antonello et al., 2021). This suggests that a key difference between neural networks is how contextual information is integrated in the word embeddings. This further suggests that contextual information might be a key feature to explain brain activity in different brain areas.

### c. Partition of linguistic representations along the gradient

A linguistically motivated processing hierarchy assumes increasingly abstract representations of language input, spanning phonemes, words, phrases and full narratives. Such a hierarchy emphasises the qualitative differences between what type of information is being integrated and is orthogonal to the previously discussed effect of context. In fact, context can not only modulate representations at all linguistic levels but also lead to interactions between levels. For example, the probability of a word occurring given its context also shapes predictions about upcoming phonemes (Heilbron et al., 2020). Different levels of abstraction, as defined by linguistic theory, have been shown to be characterised by dissociable neural signatures. To reveal such feature-specific neural markers, a language model's predictions can be used as tools to approximate a given feature's expectedness (Donhauser & Baillet, 2020; Heilbron et al., 2020).

A more direct approach to probe the information encoded in hidden representations of artificial neural networks has been taken in the field of natural language processing. NLP researchers have applied a suite of tools, such as the analysis of a model's predictions to specific, carefully controlled sentences, linear decoders or probes, representational similarity analysis and model ablation. These approaches provide evidence that artificial neural networks trained on language tasks develop a rich set of both semantic and syntactic knowledge (Rogers, Kovaleva, & Rumshisky, 2020; Wattenberg, Viegas, & Coenen, 2019), including hierarchical syntactic representations (Manning, Clark, Hewitt, Khandelwal, & Levy, 2020), subject-predicate agreement (Gulordava, Aina, & Boleda, 2018), syntactic and semantic roles, as well as semantic relations.

Most recent model architectures consist of multiple layers of units, with transformations being implemented in the connections between layers as well as in additional attention heads. If a hierarchical structure of representations should emerge from these models it should therefore necessarily be constrained by their intrinsic architecture. On aggregate measures, *i.e.* across multiple sentences, it appears that different types of linguistic knowledge can be partly localised to specific network layers or groups thereof. Specifically, information regarding syntactic structure seems to interact with degree of context along a hierarchy of layers, such that deeper models are able to decode deeper parsing tree layers (Blevins, Levy, & Zettlemoyer, 2018) and more high-level combinatorial information, such as coreference (Tenney, Das, & Pavlick, 2019). However, on a sentence-by-sentence basis, it can be noted that any hierarchical order of layers is dynamically adjusted within transformer models. For example, higher-level representations, such as semantic roles, may emerge earlier in the hierarchy if they are needed to disambiguate information at lower levels, such as part-of-speech (Tenney et al., 2019). This flexibility in how layers can temporarily reorganise the processing hierarchy essentially replaces the need for feed-back connections and is possible due to transformer models selecting relevant context through self-attention, taking into account both previous as well as following words.

Several groups have explored whether representations emerging at different layers of the language models vary in their capacity to map onto brain data. One emerging pattern across studies is that middle layers map better onto brain activity as compared to early or late output layers (Anderson et al., 2021; Caucheteux & King, 2022; Kell, Yamins, Shook, Norman-Haignere, & McDermott, 2018; Schrimpf et al., 2021; Thompson, Bengio, Formisano, & Schönwiesner, 2021; Toneva & Wehbe, 2019; Wehbe, Vaswani, et al., 2014). This is interesting, given the machine learning results reviewed above, which suggest that middle and late layers encode the most abstract linguistic information. Since these layers are closest to the model output layer, they preserve the information that is most useful for the downstream linguistic task the model is trained on (Rogers et al., 2020). This could also suggest that humans and artificial neural networks systematically differ in the last layers because their computational goal is different. Indeed, on one side, artificial neural networks predict the upcoming word using only information present in the text. On the other side, humans supposedly infer meaning by further integrating other sources of information coming from the external world through perception and action. It should be noted, however, that differences in encoding performance across layers are generally small. For example, Anderson and colleagues report only a small drop from 76% to 70% rank performance score (50% signalling chance performance) in predicting brain activity when using different layers of BERT (Anderson et al., 2021).

### d. Changes in representational geometry along the gradient

Representations emerging in artificial neural networks are highly structured. While some of this structure is driven by the architectural configuration of layers as discussed above, some structure also emerges through learning and is due to task demands both within layer-specific embeddings as well as within additional attention heads. The latter is not always explicitly taken into account when using those structured representations to model brain data.

One meaningful low-dimensional vector representation of a set of stimuli is one in which those stimuli that belong to a common abstract category will cluster together in the embedding space. Such clusters, also referred to as manifolds, usually span only a subset of the full embedding space, *i.e.* they are low-dimensional. Neural manifolds take on different shapes and dimensionalities in different language models, but some degree of clustering is observed across all (Cai, Huang, Bian, & Church, 2020). For example, in the context of supervised learning of visual classification tasks, artificial neural networks optimise the shape, dimensionality and distance of manifolds such that task-relevant abstract classes they represent become linearly separable to the best degree possible (Cohen, Chung, Lee, & Sompolinsky, 2020). Similarly, a gradual compression of word manifolds can also be observed across layers in trained artificial neural networks that process natural language. Increased linear separability across layers has been reported for abstract classes like part-of-speech, entities and dependency depth, though, notably, these increases were restricted to items with ambiguous word sense (Mamou et al., 2020). The concept of manifolds has great potential because it describes geometric objects that can be studied with precise mathematical tools. Characterising the manifold geometry has inspired a range of metrics that quantify computationally relevant changes to neural representations. Data-driven approaches such as clustering and dimensionality reduction (e.g. decomposition by non-negative matrix factorisation or PCA) are already being used to gain deeper insights into the brain's functional topology (Hamilton, Edwards, & Chang, 2018; Schoffelen et al., 2017). Leveraging the concept of linear separability of manifolds, mean field capacity analysis has been developed to quantify subtle characteristics of the representational manifold such as its radius, dimensionality and centre-correlation (Cohen et al., 2020).

Furthermore, a separation between semantic features and relational encoding has been observed in trained language models. For example, within the low-dimensional context embeddings, semantic and syntactic information begin to occupy orthogonal subspaces (Wattenberg et al., 2019). In transformer models, the introduction of the attention heads and a separate position encoding for each word further reinforces such a separation. This allows the model to learn flexible projections between relational information and semantic features. Indeed, in fully trained models, at least a subset of the attention heads seem to encode highly focused dependency relations such as determiner-noun or object-verb relations (A. Clark, 2013; K. Clark, Khandelwal, Levy, & Manning, 2019; Manning et al., 2020; Wattenberg et al., 2019). Because information encoded across attention heads is more localised (a single dependency relation is usually represented by just one attention head) using attention heads as feature models can reveal a more fine-grained functional mapping across brain areas (S. Kumar et al., 2022). The computational principles that underlie this factorisation of knowledge in transformer models have been linked to those in models of hippocampal functioning (Whittington, Warren, & Behrens, 2021) which learn a spatial code that is separate from the representations of object identity in the environment. A similar point about separating relational codes from feature codes has been made regarding the division of labour of ventral and dorsal streams in scene perception. Specifically, low-dimensional manifolds representing relations between objects in a scene might serve the purpose of enabling transfer of abstract relational information (Summerfield, Luyckx, & Sheahan, 2020).

Close inspection of the representational geometry within artificial neural networks allows us to extract more specific hypotheses about the underlying computational principles (Jazayeri & Ostojic, 2021). Having established that sentence embeddings captured in current language models broadly map onto neural representations in the brain, we can now constrain this mapping by explicitly taking into account the representational geometry evident in those language models. Different stimulus materials might drive clustering of neural representations in slightly different ways, possibly leading to the discrepancies in identifying higher-level "high-context" brain areas. The reported redundancy in the neural code might be a consequence of different brain areas encoding semantic information that corresponds to distinct subspaces of the model embedding vector.

# IV. Future directions towards interpretable process models of sentence comprehension in the human brain.

The development of artificial neural networks to process natural language has two advantages. First, they are trained end-to-end and thus provide objective measures that do not depend on a priori assumptions concerning how language is structured and how humans are born with in-built knowledge of this structure. Second, they can process natural language and allow the analysis of brain data in the context of natural language comprehension. However, it remains unclear to what extent these models are to be taken as process models of the brain, *i.e.* to what extent they align on an algorithmic level to the brain and how well the computations that they realise are similar to the computations that the brain realises (Barsalou, 2017). We identify three challenges that need to be solved to make these artificial neural networks good candidates as true process models of natural language comprehension in the brain.

### a. Improving interpretability

Artificial neural networks are sometimes depicted as "black box models", because it is hard to describe how they work at an algorithmic level. However, some modelling approaches are harder to interpret than others. For example, word embeddings that directly rely on corpus co-occurrence statistics allow a direct interpretation of the coefficients of the regression, because each dimension transparently represents the co-occurrence with one other word in the vocabulary (Huth et al., 2016; Popham et al., 2021). For example, a "*river*"-specific voxel would have a large weight for "*river*" and a low weight for all other dimensions. On the contrary, learned embeddings such as GPT-2 have arbitrary dimensions that are not directly interpretable. Having a more transparent mapping between the algorithmic level and the input stimulus features does not guarantee high interpretability. In fact, it has been argued that due to the high-dimensional parameter space of complex neural systems the emerging solutions likely rely on very complex nonlinear interactions between multiple features which might not neatly fit onto human-interpretable dimensions (Hasson, Nastase, & Goldstein, 2020). Nonetheless, having a transparent mapping from network representations to stimulus features seems to be a prerequisite for identifying and mapping the emerging algorithmic properties onto our existing cognitive theories as well as brain networks.

On the methods side, some approaches result in more or less interpretable results. Comparing brain activity with the model's using linear regression is the most direct test but it yields little insight about how the brain works if the model itself is uninterpretable. Alternatively, using metrics derived from the model's output, such as surprise, can be more interpretable. For example, Heilbron and collaborators computed interpretable metrics from the GPT-2 activations, namely the phonemic, syntactic and semantic surprise associated with each word (Heilbron et al., 2020). The linear regression of these metrics on brain activity yields interpretable results, *e.g.* here the fact that the brain is processing phonemic, syntactic and semantic aspects in a hierarchical way, at different times and in different locations.

Another way to improve the interpretability of the results is to analyse the model's activity prior to using it to fit brain activity. This involves having a better qualitative understanding of the different clusters of artificial neurons that compose the model. For example, Caucheteux and collaborators (Caucheteux & Gramfort, 2021) clustered and analysed the GPT-2 activations to disentangle syntactic from semantic composition. They were then able to localise using fMRI the brain regions selectively associated with syntactic and semantic composition, beyond the mere representation of the individual syntactic or semantic features of each word. We believe that the field should rely on this general approach more heavily and could benefit from the tools that have been developed in the machine learning community to probe representations within artificial neural networks, such as linear decoders, representational similarity analysis and representational geometry analysis. Identifying the subcomponents that compose the artificial neural networks and their specific role will improve the interpretability and the insight gained from comparing representations between models and the brain. Previously, we discussed the effect of shallow versus deep context in artificial network representations. Beyond quantifying the number of words in the context, we are currently lacking a more qualitative description of what sort of information might co-vary with context depth. For example, an intuitive distinction is the difference between adjacent and nonadjacent dependency relations. Can we identify further clusters apparent in artificial neural network representations that are modulated by contextual depth but not routinely taken into account in current process models?

Finally, a detailed algorithmic understanding might only be useful if a mapping onto the brain can be established later on. Therefore, we do not want to argue that algorithmic interpretability should be the sole focus going forward. The complementary objective of improving the model-to-brain mapping will also be required. Some attempts at optimising a neural network for its similarity to the brain have reported increased task performance in both object recognition and NLP tasks (Kubilius et al., 2019; Toneva & Wehbe, 2019).

  b. **Controlling for the relative contribution of the model and the linear regression to fit of the brain activity**

The activity of any large language model typically contains a lot of information about the stimuli themselves. As the brain is responding to those stimuli, the models might fit the brain simply because the stimuli themselves are helpful to predict brain activity. Indeed, there are two transformations playing a role in the fit: the non-linear encoding of the stimuli by the model and the (linear) regression between the model and the brain. A large part of the predictive power of these models could actually be due to the linear regression between the models and the brain activity. Indeed, with a sufficient number of parameters, the linear regression can learn the average brain activity of each trial. One way of measuring the relative contribution of the model and the linear regression is to use artificial neural networks with random weights. This equalises the number of parameters fed to the regression and the information about the stimuli while removing what has been learned by the model during training. Several studies have run this control on fMRI data recorded during language processing (Kell et al., 2018; Millet & King, 2021; Millet et al., 2022; Schrimpf et al., 2021). They confirm that trained models are better than untrained ones to explain brain activity. This proves that the predictive power of these models is at least partly related to the way they actually process the stimuli. However, there are two major caveats. First, the added value of trained models over untrained ones is usually relatively small, ranging from less than 1% to 53% in (Schrimpf et al., 2021). Second, it has been reported that random models with the same architecture but a different number of layers can yield to large differences in predictive power, from 0.2 to 0.6 between different versions of GPT (Schrimpf et al., 2021). This calls for caution when interpreting the results because it suggests that at least part of the predictive power of these models is explained by the linear regression and not necessary because they are good process models of the human brain.

### c. Situation model

One last issue is how to place language comprehension in an integrated comprehension system. Indeed, the artificial neural networks used as models of language comprehension in the brain focus exclusively on language-internal tasks, *e.g.* next-word prediction tasks. These models never observe the situation in which the language is produced, and have no knowledge of the locutor, much less of any covert intentions. This is different from how humans process language.

First, it is known that the situation model plays an important role in language comprehension. For example, it is almost always the case that language happens in a communicative context, where several streams of information in addition to the speech signal itself have to be taken into account. For example, pragmatic knowledge has to interface with the semantic information provided by the words. The idea of separating pure semantics from pragmatics has been heavily criticised, for example by Jackendoff (Jackendoff, 2003). Another example are visual cues such as gestures that accompany speech production. Indeed, it is known that gestures provide semantic information that is integrated with the speech information to form a coherent representation (Kelly, Ozyürek, & Maris, 2010; Vigliocco, Perniss, & Vinson, 2014; Willems, Ozyürek, & Hagoort, 2007). More generally speaking, the current artificial networks only exploit the statistics of the texts, whereas another critical source of information is the statistics of the world. Indeed, sentence production and comprehension co-occurs with multiple sources of information, that includes all kinds of sensory information, such as non-verbal, visual and auditory percepts, so that word meaning is actually grounded in the external world. Such factors have not been taken into account so far by the artificial neural network modelling approach. However, it should be noted that there are in principle no limitations to do so. Indeed, gestures can be fed as additional input features to the artificial neural networks. As such, they would provide contextual information that could be used by the artificial neural networks to produce better contextualised embeddings of the words.

Second, another issue concerns the computational goal of the artificial neural networks that process natural language, *i.e.* their objective function. Language is used by humans to convey information about the world whereas the sole goal of the artificial neural networks is to predict masked words using the surrounding words. When presented with a descriptive text or speech, a human listener constructs a representation of the described situation (Zwaan & Radvansky, 1998). This representation helps support comprehension and helps drawing inferences about the situation. Take the example introduced by McClelland and collaborators (2020): "*John spread jam on some bread. The knife had been dipped in poison.*". A human listener might infer from this sentence that the jam was spread with a knife, that poison has been transferred to the bread, and that if John eats it, he may die (McClelland, Hill, Rudolph, Baldridge, & Schütze, 2020). This inference is based on a rich representation which incorporates prior world knowledge. It is unclear how to incorporate such knowledge in artificial neural networks. At the moment, these networks are mostly trained on masked word prediction tasks and are not forced to draw inferences on the external world. Recent attempts include incorporating a long-term memory integrated representational system (McClelland et al., 2020), or training language models to recognize or even generate images (Ramesh, Pavlov, Goh, & Gray, 2021). It should nonetheless be noted that the word embedding produced by the artificial neural networks actually aligns to common human knowledge, such as information about object features like size or shape (Grand, Blank, Pereira, & Fedorenko, 2022).

# Conclusion

Artificial neural networks are promising tools to study brain activity during sentence comprehension. They generate explicit instantiations of low-dimensional vector-based representations of words. Once a model is sufficiently trained on a language corpus, these representations can be extracted automatically and quickly from large natural language stimulus datasets. They therefore facilitate the analysis of neuroimaging data recorded during natural language comprehension of sentences and narratives. Their success in predicting brain data suggests that these representations can indeed be useful models of sentence-level meaning representations. Furthermore, the representations generated by the most recent language models are modulated by features relevant to sentence processing, such as context depth.

Nonetheless, the insights into the functional hierarchy of the brain network for language that deep learning models provide are currently very coarse, which makes it difficult to complement current hypotheses about cognitive process models. The main limitations are their interpretability and their ability to be good process models of the brain. In particular, most models are hard to interpret, they consist of word embedding with arbitrary dimensions that do not map onto simple concepts or simple syntactic roles. Furthermore, most models are trained on next-word prediction tasks, without any reference to the external world and without a communicative function. It is also currently unclear to what extent the predictive power is explained by the regression only and not so much by the ability of the model to capture brain processes.

In order to make the most out of these models, we suggest that future directions could involve: (1) more constrained tests when directly comparing models activation and brain activity using regression methods, (2) better descriptions of the model's internal representations, *i.e.* what information is clustered and what information is factorised. One good starting point would be to clarify which aspects of a sentence's meaning is modulated by context, beyond a binary dichotomy between single words and words in context. Here, language researchers can borrow techniques from the machine learning literature on deep learning models, where multiple analysis techniques have been explored to provide insight into the inner workings of recent deep language models.